# Anomaly Detection for Unmanned Aerial Vehicle Sensor Data Using a Stacked Recurrent Autoencoder Method with Dynamic Thresholding

**Victoria Bell[1], Divish Rengasamy [2], Benjamin Rothwell [2], * and Grazziela P Figueredo [3]**

[1] School of Computer Science, The University of Nottingham, Nottingham, NG8 1BB, UK; Victoria.Bell2@nottingham.ac.uk
[2] Gas Turbine and Transmissions Research Centre, The University of Nottingham, Nottingham, NG7 2RD, UK; divish.rengasamy@nottingham.ac.uk
[3] The Advanced Data Analysis Centre, The University of Nottingham, Nottingham NG8 1BB, UK; grazziela.figueredo@nottingham.ac.uk
* Correspondence: benjamin.rothwell@nottingham.ac.uk

**Abstract:** With substantial recent developments in aviation technologies, Unmanned Aerial Vehicles (UAVs) are becoming increasingly integrated in commercial and military operations internationally. Research into the applications of aircraft data is essential in improving safety, reducing operational costs, and developing the next frontier of aerial technology. Having an outlier detection system that can accurately identify anomalous behaviour in aircraft is crucial for these reasons. This paper proposes a system incorporating a Long Short-Term Memory (LSTM) Deep Learning Autoencoder based method with a novel dynamic thresholding algorithm and weighted loss function for anomaly detection of a UAV dataset, in order to contribute to the ongoing efforts that leverage innovations in machine learning and data analysis within the aviation industry. The dynamic thresholding and weighted loss functions showed promising improvements to the standard static thresholding method, both in accuracy-related performance metrics and in speed of true fault detection.

## 1. Introduction

Autonomous aircraft need to be able to safely navigate through largely dynamic environmental conditions, making anomaly detection during UAV flights a crucial area of study. The quality of the system in-use determines both the safety of the aircraft, and the monetary cost of maintenance. Moreover, as newer UAV technologies are explored, detecting anomalous behaviour and defining their causes is essential in developing the next frontier of unmanned aircrafts. The main aim of this project is to design, develop and validate a system capable of accurately detecting anomalies in UAV benchmark data. This system must both be able to differentiate between anomalous data, and non-anomalous data, and therefore, the performance is measured using true positive and true negative classification results. Furthermore, as with anomaly detection in time-conscious domains, early and accurate fault classification can greatly reduce the impact that the outliers have on the system, since damage mitigation action can be taken. Current popular approaches for anomaly detection for UAVs focus on recurrent regression methods, in order to utilise contextual information to influence anomaly classification. There is, however, limited research into the harbouring of contextual information during the thresholding process. This paper investigates the application of a Long Short-Term Memory (LSTM) deep learning autoencoder based method and proposes a novel dynamic thresholding algorithm combined with a weighted loss function to improve the accuracy and reduce the delay when detecting anomalies on a UAV sensor log dataset. The data set put forward by Keipour *et al* [1], ALFA, will be used for this study.

## 2. Background and Related Work

LSTM prediction models aim to predict the values of data $x_T$, where T is the time of recording a datapoint x, by utilising values of $x_{T-1}$, $x_{T-2}$, … $x_{T-n}$. An LSTM unit is commonly composed of a cell, an input gate, an ouput gate, and a forget gate. These three gates regulate the flow of information in and out of the units memory. As a result, the unit can store important short-term information for an abritrary time interval to help inform future predictions. First proposed by Gers *et al* in 1999 [2], the arcitecture has been widely used and adapted in a variety machine learning application domains with concern to sequential and time-series data [3]. This time-series approach can be particularly effective when detecting anomalous behaviour in aircrafts, and other types of physical machinery, since it provides context to the data, allowing for a higher-level of circumstantial information to be deduced. LSTM predicts the values of a datapoint and compares it with the input value of the instance. If the disparity between the prediction and true values exceeds a specified threshold, then the datapoint is classified as anomalous. An LSTM-based method has been found to be greatly effective when applied to one-dimensional UAV sensor data, with a total of 997 datapoints test, all 30 anomalous data points were correctly identified, with only 1 false positive classification [4]. A number of variations of LSTM neural networks have been proposed and applied to UAV sensor data, providing promising results. For example, Wang *et al* [5] produced an LSTM-based method with residual filtering (LSTM-RF), which was applied to both real UAV flight data, with simulated "faulty" anomalous instances. The residual filtering method aimed to smooth the residuals in order to promoto fault detection performance. This method was compared to LSTM without residual filtering [6], and a least-squares support vector machine (LS-SVM) [7]. The LS-SVM model was also supplemented with residual filtering. The results revealed that the introduction of residual filtering successfully reduced the noise when making regressions used during the fault detection process. As a result, the proposed LSTM-RF outperformed LSTM on the real data, achieving an increase of 0.451 and 0.133 of the accuracy scores of both the fault types tested. This paper demonstrated a considerable positive effect on anomaly detection performance of an LSTM architecture by a post-inference residual manipluation. Similarly, this paper investigates a post-inference statisitcal technique to enhance the performance of LSTM architecture for fault detection on UAV data.

An autoencoder is an artificial neural network that processes data with the goal of encoding and decoding (reconstructing) the data with minimal reconstruction loss. In order words, the goal is to minimise the loss between the input of the model, and the output. During the encoding stage, the transformation algorithm applied to the data reduces the dimensionality. The decoding stage aims to return the data into its original form, with the original dimensionality. . A simple example of an autoencoder is displayed in Figure I:

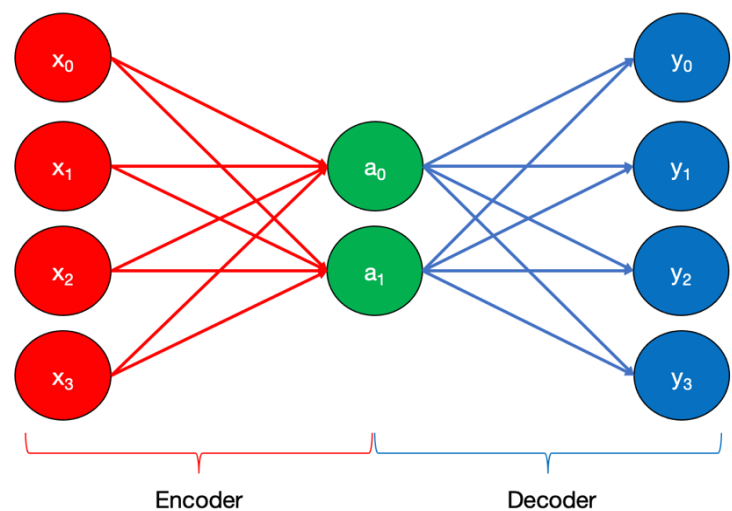

*Figure I: Architecture of a simple autoencoder, encoding data with a feature length of 4 ($x_0, \ldots x_3$) into 2 features ($a_0, a_1$), and then decoding the hidden layer back into 4 features ($y_0, \ldots y_3$).*

The training of autoencoders are unsupervised, and therefore have the advantage of being well applicable to domains where anomalous data may be largely heterogenous. Research into the use of autoencoders for UAV sensor data is relatively limited; however recent interest has been generated in the area, as well as their applications of fault detection of autonomous machines in general. An incrementally trained variational autoencoder method (Incr-VAE) encorporating Bidirectional LSTM (Bi-LSTM) layers, when applied to autonomous robot system logs, was found to obtain the best anomaly detection performance when compared to a number of other methods [8]. These methods included a convolutional (2d) autoencoder and a state-of-the-art LSTM Variational Autoencoder (LSTM-VAE). One study that is referred to throughout this work, was work done by Park *et al* in 2021, where an autoencoder was applied to the same ALFA dataset as detailed in Section 3.1, for the purpose of outlier detection [9]. The study investigated the performance of a stacked autoencoder at a variety of threshold levels, and the results exhibited high potential for this method, obtaining strong AUC scores for each type of failure.

Park *et al* however focused on a significantly smaller portion of the data. Our work aims at overcoming this limitation by including all data available. Park *et al* results are therefore not suitable as a direct benchmarking comparison to our work, and instead serve as a performance reference.

When applied to the task of anomaly detection, the output of an autoencoder is compared with the original data. In this paper, the term "thresholding" refers to the selection of a value used for this comparison when making a classification assertion (e.g., classifying an instance as anomalous or non-anomalous). Static thresholding, or parametric thresholding is frequently used in anomaly detection for vehicle sensor data. As vehicle technologies have increased greatly complexity, it has been necessary to implement thresholding systems that can extract a better behavioural image of the data. As a result, the thresholding technique should account for the intricate relationships between different features, such that more niche anomalies that could not be identified by direct human analysis can be identified. Research performed at the Nasa Jet Propulsion Laboratory implemented a dynamic thresholding approach to spacecraft data and compared it with the traditional static thresholding approach in the task of anomaly detection [10]. The goal of this study was to reduce the frequency of false positive results. The results found that dynamic thresholding decreased the recall percentage points score by 4.8 however successfully increased the overall precision by 38.6 percentage points [11]. The combination of a dynamic thresholding strategy and LSTMs, both of which make use of contextual data, is investigated in this paper as an improvement of anomaly detection on time-series UAV data.

Rengasamy *et. al* [12] proposed a dynamically weighted loss function for the purpose of forcing deep learning models to prioritise learning training instances with high relative losses. This method was applied to two case studies: (1) gas turbine engine remaining useful life (RUL) using commercial modular aero-propulsion system simulation (CMAPPS), and (2) air pressured system (APS) fault detection in trucks. The weighted loss function was found to improve the performance in both case studies, and most noteably with relation to this paper, the inclusion of the weighted loss function sucessfully reduced the Scoring Function results for the by 27.7% and RMSE by 30.6% when applied to a bidirectional LSTM neural network for the CMAPPS dataset.

### 3. Methodology

#### 3.1 Data Collection/Description

The dataset used in this paper was obtained from the Carnegie Mellon University website [13]. The data was collected by Keipour *et al* [14] across a total of 47 autonomous flights of a fixed-wing Carbon-Z T-28 UAV (**Error! Reference source not found.**) with an onboard computer. The original firmware was altered such that faults could be synthetically produced by disabling parts of the aircraft during flights. Sensors recorded over 350 variables during the flights. The plane has a 2-meter wingspan, a single electric engine at the front of the craft, ailerons, flaperons, an elevator, and a rudder. The plane was equipped with a Holybro PX4 2.4.6. autopilot, a Pitot Tube, a GPS module, and a Nvidia Jetson TX2 for data logging.

During flight, the UAV is completely autonomous, unless aircraft safety is compromised, in which a pilot can take control remotely from the ground control station. Moreover, the ground control station is used for disabling control surfaces for instigating synthetic faults.

The 47 flights comprise of 10 non-faulty flights, and 37 flights with four general types of faults: engine, aileron, rudder, and elevator. Each flight is independently logged, with the known time of failure being provided, as well as a fault detection log, with the relevant timestamps. Each feature is recorded at rates ranging from 1Hz to 50Hz. The faulty flights have an additional feature, failure status, that indicates the timestamps that contain anomalous data. Out of the 37 faulty flights, 23 are full engine power failures, 3 are due to rudder failures, 2 are elevator failures, 8 are aileron failures, and 1 is a combination of rudder and aileron failures (as shown below in Table 1).

*Table 1: Description of the full ALFA dataset*

| Fault Type | Number of Flights | Time Before fault (s) | Time after Fault (s) |
|---|---|---|---|
| Engine full power loss | 23 | 2,282 | 362 |
| Rudder stuck to left | 1 | 60 | 9 |
| Rudder stuck to right | 2 | 107 | 32 |
| Elevator stuck at zero | 2 | 107 | 23 |
| Left aileron stuck at zero | 3 | 228 | 183 |
| Right aileron stuck at zero | 4 | 442 | 231 |
| Both ailerons stuck at zero | 1 | 66 | 36 |
| Rudder and aileron at zero | 1 | 116 | 27 |
| No fault | 10 | 558 | - |
| **TOTAL** | 47 | 3,935 | 777 |

**3.2 Timestamp Pooling**

As mentioned in 3.1, the UAV sensors independently log feature values at rates ranging from 1Hz to 50Hz. Since the stacked autoencoder architecture, like most machine learning methods, are not directly compatible with this format, as they require a set number of features per data point, it is necessary to perform timestamp pooling. In this experiment, prior to analyzing the data, a timeframe of 200 milliseconds was chosen. At each 100-

millisecond increment, a new data point would be formed by the values of the features closest to this point in time, as in Figure II:

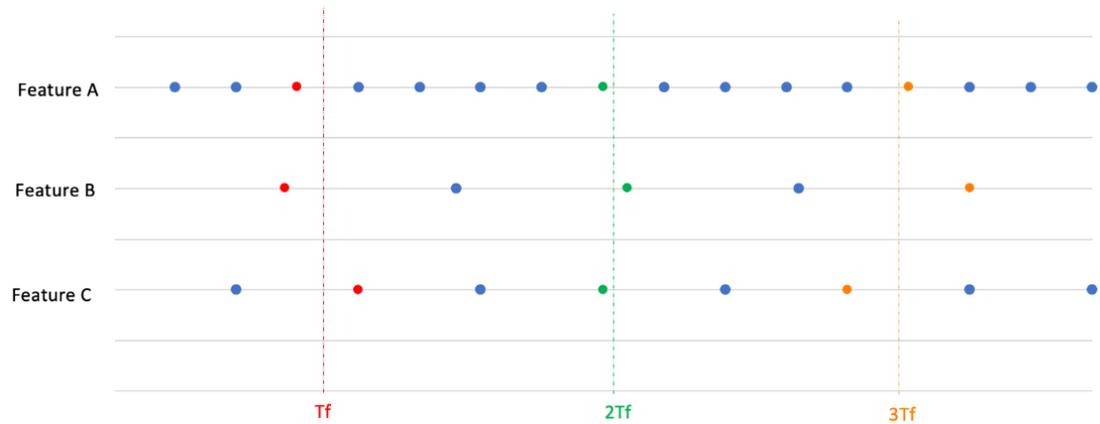

*Figure II – A diagram to demonstrate the process of timestamp pooling, with Tf, 2Tf, 3Tf representing the timestamps of different intervals, with the colours co-ordinated with the datapoints selected from features A-C to form the new dataset.*

This method was deemed more appropriate than the alternative of choosing a random value within the timeframe since it was essential to form data as realistically as possible, such that the identifying characteristics of anomalous data would not be lost. However, this solution is not faultless since the values chosen for each set of feature values will be somewhat deviating from their actual values at each real timestamp. To reduce the severity of this problem, 100 milliseconds is also the maximum limit time difference during value selection. Therefore, if a particular feature does not have a value inside 200-millisecond thresholds for a timestamp, then the whole timestamp is discarded. Through data analysis and trial-and-error, a timeframe of 100 milliseconds was chosen since this value gave a suitable balance of data accuracy while limiting the number of data points that needed to be rejected.

**3.3 Feature Selection and Normalisation**

The raw ALFA dataset contains more than 350 features, many of which are not relevant to the purpose of outlier identification, and the inclusion of which would add noise and reduce model performance. The first stage employed for feature selection involved manually inspecting the dataset and identifying features that are to be rejected from the experiment. The rejection of a feature would be determined via two rules:

1) The feature will be discarded if it is a derived feature (e.g., a measure of covariance between two logged features)

2) The feature will be discarded if the sensor stability is inadequate (e.g., the feature does not contain reliable log sources that change throughout the flight and does not contain null values).

The remaining 120 features are then normalized using the Z-Score standardisation. Z-Score, or standard normalisation, transform the data such that it has the same properties of a normal standard distribution, with a mean equal to 0 and a standard deviation of 1. The equation for Z-Score normalisation is described as:

$$Z(x) = \frac{x - \mu}{\sigma} \qquad (1)$$

Following normalisation, a linear Principal Component Analysis (PCA) is applied to reduce the dimensionality further. PCA is a non-parametric statistical technique applied to a dataset in order to form a set of uncorrelated features that can effectively represent the dataset at a lower dimensional space. In this paper the dataset is reduced via PCA to 50 features, accounting for 99.9% of the 120-feature dataset variance.

**3.4 LSTM Stacked Autoencoder**

During training, autoencoders aim to determine the most informative features of the input data, such that minimal information about the values of other features is not lost during dimensionality reduction. A stacked autoencoder is a combination of multiple sub-autoencoders, whose outputs are streamed as input into the sequential sub-autoencoder. In this paper, 3 LSTM sub-autoencoders are utilised, reducing the dimensionality from 50, to 30, to 15, and to 10 in sequence (following a reconstruction). The more dimensionality is reduced, the more the autoencoder is forced to establish the most informative features, and in outlier detection, this is a valuable tool in identifying anomalous behaviour. Using a singluar autoencoder to severely reduce dimensionality comes with risks of high

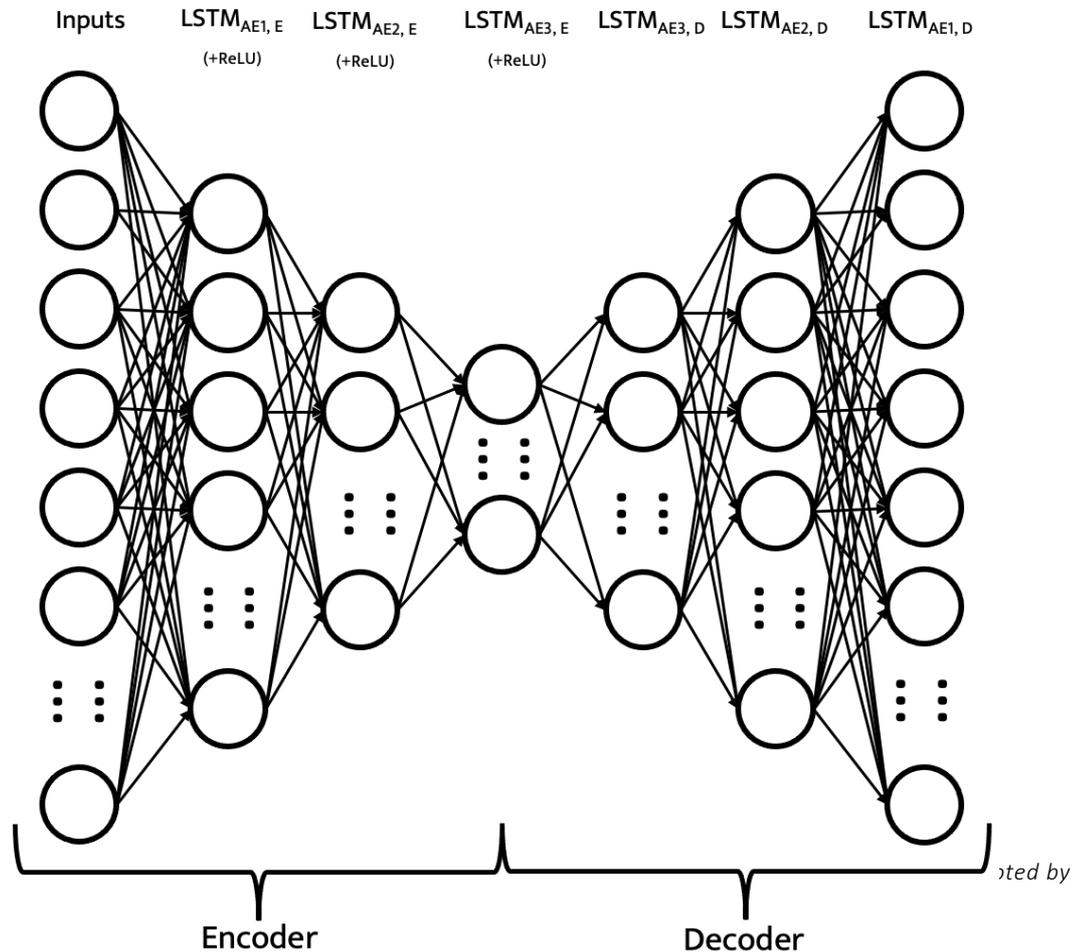

reconstruction losses, since there would be large spatial difference between the raw data and encoded data, and the autoencoder would simply not be able to represent all the relevent information. Stacking the autoencoders enables the model to reduce dimensionality stepwise, thereby smoothing the slope, and retaining more relevent information, at a low dimensional-space. A diagram of the LSTM stacked autoencoder architecure is displayed below in Figure III:

In this paper, during the encoding phase, a single-layer LSTM function (Equations 1-6) is performed on the input features that reduces the dimensionality, followed by a non-linear ReLU (rectified linear unit) function. During the LSTM layer function, the hidden state at time $t$ ($h_t$) is computed as follows:

$$i_t = \sigma(W_{ii}x_t + b_{ii} + W_{hi}h_{t-1} + b_{hi}) \qquad (2)$$

$$f_t = \sigma(W_{if}x_t + b_{if} + W_{hf}h_{t-1} + b_{hf}) \qquad (3)$$

$$g_t = tanh(W_{ig}x_t + b_{ig} + W_{hg}h_{t-1} + b_{hg}) \qquad (4)$$

$$o_t = \sigma(W_{io}x_t + b_{io} + W_{ho}h_{t-1} + b_{ho}) \qquad (5)$$

$$c_t = f_t \odot c_{t-1} + i_t \odot g_t \qquad (6)$$

$$h_t = o_t \odot tanh(c_t) \qquad (7)$$

Where $i_t$, $f_t$, and $o_t$ are the the input, forget and output gate states at time t, $g_t$ and $c_t$ signify the cell state values at time t, $\sigma$ denotes the sigmoid function, and the symbol $\odot$ denotes the element-wise product of two matrices.

Via the LSTM layers, the hidden state is activated by the input states at previous timestamps. This allows for the model to incorporate contextual information when formulating a prediction. The ReLU activation function only deactivates the neurons whose output of the LSTM layer is less than 0, thereby reducing the computation cost of the calculations and reducing the number of iterations required for the model to converge. Moreover, ReLU introduces sparse respresentation of the features learned, and as a result, in theory, generalises more effectively. The Relu function is defined by the following equation:

$$ReLU = (x)^+ = max(0, x) \qquad (8)$$

During the decoding stage, a second-layer LSTM function is applied to the data of a lower dimensional latent space, to return it to its original dimensionality. ReLU is not applied at this stage, since some of the reconstructed features have negative values post-normalisation.

**3.5 Loss Calculation**

The fundamental loss values are calculated between the true data values used as input and the reconstructed data values using the mean squared error (MSE) function, as defined:

$$MSE(x_i, y_i) = \frac{1}{n}\sum_{i=1}^{n}(x_i - y_i)^2 \qquad (9)$$

In conjunction with the basic loss calculation, a dynamically weighted loss function is also applied. The algorithm employed was first proposed by Rengasamy *et al*, with the purpose of encouraging the model to focus on learning the more marginal data instances, e.g., the datapoints with the higher losses. This is achieved by multiplying the result of the basic loss calculation by a weight variable D:

$$L(x_i, x_r) = D * MSE(x_i, x_r) \qquad (10)$$

The value of D is determined by the performance of the model during model prediction. By considering instances with an absolute loss value of less than (or equal to) a constant threshold, *c*, as being successfully learnt, the value of weight for the parameter D will be low. By contrast, instances with absolute loss values of more than *c* are multiplied by a high value of D, such these instances have a more significant effect when the model is learning. The final loss of our approach is therefore defined as follows:

$$L(x_i, x_r) = D(x_i, x_r) * MSE((x_i, x_r) \qquad (11)$$

$$D(x_i, x_r) = \begin{cases} \frac{|x_r - x_i|}{2} & if\ |x_r - x_i|\ is < C \\ |x_r - x_i| & otherwise \end{cases} \qquad (12)$$

This is a particularly effective algorithm for anomaly detection with autoencoders, since the goal of training is not only to reduce the mean loss of the training data comprised of solely safe flight logs, but to reduce the deviation of performance between easily learned safe instances and the more marginal instances. As a result, we hypothesise that this should reduce the number of false positive results during the testing phase.

### 3.4 Dynamic Thresholding

Traditional autoencoders anomaly detection methods often use a static/linear thresholding approach during the anomaly classification stage. In aviation time-series data, a limitation of this method is that it does not consider the possible information that could be gathered when considering the behaviour of the aircraft prior to the instance timestamp, and the relative performance of the model during the specific stage of flight. This paper proposes a novel thresholding mechanism, that encompasses the importance of contextual anomaly classification. Since flight data is, by nature, sequential, the reconstruction loss values on previous datapoints within the flight ($x_{n-j}$, $x_{n-j+1}$, … $x_{n-1}$) to determine the thresholding value for the current data ($T_n$). This is calculated by the sum of overall mean loss of all prior datapoints and their mean absolute deviation. This calculation is described below (13-16):

$$T_n = T(x_n) = W_y L + W_z (M + S) \quad (13)$$

$$L = \mu_{train} + \sigma_{train} \quad (14)$$

$$M = \frac{1}{j} \sum_{i=n-j}^{n} x_i \quad (15)$$

$$S = \sigma(x_{n-j} \ldots x_{n-1}) \quad (16)$$

Where $W_y$ and $W_z$ represent constants for weighting L and M respectively. In this experiment they are set to 0.3 and 0.7 respectively, determined experimentally. L is equal to what can be considered a static threshold constant obtained from the training dataset losses. M is equal to the mean of the losses of the previous j instances, while S is equal to the standard deviation of said datapoints. In this experiment, j is set to 100.

This method ensures that the information from the training stage and the losses on previous datapoints within the flight are both exploited in order to determine a thresholding value every test instance.

## 4. Experimental Set-Up

### 4.1 Training Dataset

The dataset used for training comprises of the 10 pre-processed non-faulty flights provided in the ALFA dataset. Each flight undergoes the pre-processing method as described in section 3.

### 4.2 Model Optimisation

The optimiser utilised during the training stage of this experiment is Stochastic Gradient Descent (SGD). This is an iterative method for optimizing the objective function of the stacked autoencoder model, that is, minimizing the calculated loss.

### 4.2.5 Autoencoder Training

Training of the Denoising Stacked Autoencoder entails a greedy layer-wise training algorithm, followed by a fine-tuning phase, outlined by the following four steps:

1. Train the first sub-autoencoder, $A_0$, on the pre-processed data, for 5000 epochs.

2. Train the second sub-autoencoder, $A_1$, with the input being output of the trained $A_0$, for 5000 epochs.

3. Train the third sub-autoencoder, $A_2$, with the input being the output of the trained $A_1$, for 5000 epochs

4. Train the full stacked-autoencoder with a lower learning rate to fine-tune the weighting parameters for 2000 epochs.

Due to this architecture, it was imperative that when applying the weighted loss function, four distinct values of c were determined experimentally and applied to each training step (0.5, 0.12, 0.02 and 0.8 for each stage respectively). Furthermore, the number of epochs used for each training stage were also selected experimentally to ensure convergence.

**4.3 Test Datasets**

The ALFA dataset has considerably more engine fault data than the other fault types (with 23 engine fault flights, 3 rudder failure flights, 7 single aileron failure flights, 2 elevator failure flights, 1 double aileron failure flight, and 1 combination rudder/aileron failure flight). As a result, the focus of this experiment is to observe the performance of the method on the engine fault type flights. Each flight underwent the same pre-processing techniques as the training set, with the same normalisation scale that was extracted from the training set being applied to each test flight. From these flights, the true non-anomalous data points were randomly sampled such that the number of non-anomalous data points is equal to the number of anomalous data points. This sample, and the full set of anomalous data points were used to extract the performance metric values. To ensure that the samples were a suitable representation of the full flight data, boxplots were used to visually determine whether the sample should be accepted or discarded (and subsequently a new sample would be drawn).

**4.4 Model Comparisons**

Three variations of the LSTM stacked autoencoder architecture will be tested: LSTM-AE with static thresholding (ST), LSTM-AE with dynamic thresholding (DT) and LSTM-AE with dynamic thresholding and the dynamically weighted loss function (DW). The comparison between the three models will show the effect of the proposed dynamic thresholding technique, as well as the supplement of dynamic thresholding. The static thresholding value is chosen by the mean plus one unit of standard deviation of the losses achieved on the full training dataset.

**4.5 Performance Metrics**

To evaluate the effectiveness of each model, four performance metrics are used: accuracy, precision, recall and detection delay. Accuracy is an overall measurement calculated by the number of correctly classified points divided by the total number of data points tested. This metric gives an overview of the performance of the model, however the precision and recall of the model are required to give a more comprehensive description of the performance of the model with regards to different objectives. Precision is measured by dividing the number of true positive classifications divided by the total number of positive classifications. This primarily gives an indication of performance in both correctly identifying anomalies and correctly identifying non-anomalous data. Recall is then calculated by dividing the number of true positives by the total number of true anomalies (including those that were not identified). This measurement will indicate specifically how the model is treating the true anomalies, as it is an accuracy measurement for the positive class only. The detection delay, measured in seconds, is referring to the difference between the timestamps of the first true positive datapoint, and the first anomalous datapoint correctly classified by the model. Since earlier detection is preferable when detection aviation faults, this metric allows for a domain specific performance perspective.

## 5. Results and Evaluation

Table 2 displays the mean performance metric results for the three separate methods (obtained over a series of 10 experiment runs), LSTM-AE with static thresholding (ST), LSTM-AE with dynamic thresholding (DT), and LSTM-AE with dynamic thresholding and the dynamically weighted loss function (DW).

*Table 2 – Table of results*

|  | Precision | Recall | Accuracy | Detection Delay (seconds) |
|---|---|---|---|---|
| **LSTM-AE + ST** | 0.736 ±0.09 | 0.721 ±0.17 | 0.752 ±0.11 | 2.426 ±1.83 |
| **LSTM-AE + DT** | 0.770 ±0.08 | **0.747** ±0.12 | 0.809 ±0.10 | **0.504** ± 0.55 |
| **LSTM-AE + DT + DW** | **0.793** ±0.08 | 0.735 ±0.13 | **0.821** ± 0.11) | 0.522 ±0.56 |

LSTM-AE with dynamic thresholding (second line of the table above) appears to have slightly stronger mean precision, recall, and accuracy results than LSTM-AE with static thresholding (first line of the table), and moreover a significantly shorter detection delay, identifying the true faults by 1.922 seconds on average. Since the precision is slightly higher when applying dynamic thresholding, indicating that the decrease in detection delay was not at the expense of a higher frequency of false positives. The high standard deviation of the detection delays, imply that the high delay may have been caused by specific flights, in which the model failed to classify the earlier true positive instances.

As shown in Figure IV and Figure V, when inspecting these specific flights, it is apparent the static thresholding value was entirely unsuitable to be applied to the flight and lacked the contextual information that the dynamic thresholding model employed. When dynamic thresholding is supplemented by dynamic weighted loss functions, increased in mean precision and accuracy results are obtained, while the average recall worsened slightly. Moreover, it was found that the addition of DW slightly increased the average detection delay, however this is a minor change (+3.57%). With dynamic weighted loss function, a shift in preference towards correctly classifying non-anomalous data is shown via the increase in precision score, at the slight expense of the recall. This is likely occurring since during training, the weighted loss function punishes non-anomalous values with high losses more, making the model more tolerant to these datapoints that deviate more from the expected non-anomalous behavior. During testing, this may have caused for some incorrectly classified marginal anomalous instances. As a result, however the accuracy was increased by a larger margin. Depending on the desired application of the model (e.g., whether fewer false positives or fewer false negatives are preferred), both the LSTM-DT with and without dynamic weighted loss function have their advantages.

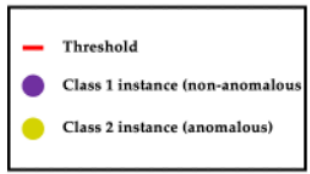
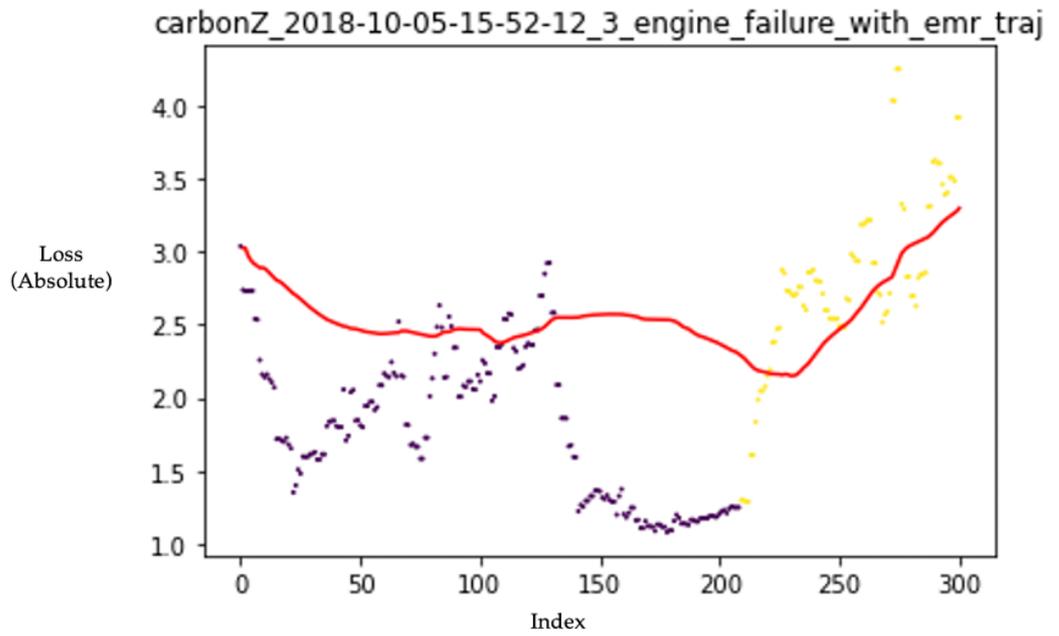

*Figure V - A diagram to represent the LSTM-AE + DT classification threshold dividing the loss of a sample engine failure flight*

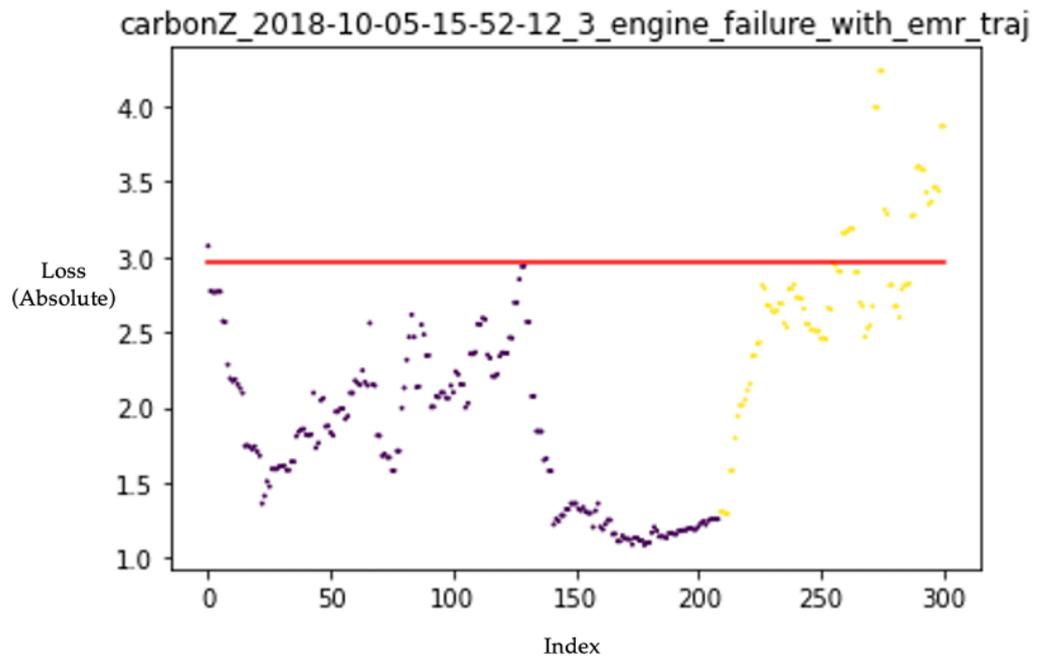

*Figure IV - A diagram to represent the LSTM-AE + ST classification threshold dividing the loss of a sample engine failure flight*

## 6. Conclusion

This paper proposed Stacked LSTM Autoencoder-based anomaly detection method combined with a novel dynamic thresholding and a weight loss function, to be applied to Unmanned Aerial Vehicle sensor data. We tested the efficacy of the approach in comparison to statistic thresholding and without the dynamic thresholding, and we found that this method displayed a strong performance in detecting anomalous data during an engine-failure scenario, with a considerable reduction in delay of true anomalous behaviour detection. The devised method is not restricted to the subject vehicle of the ALFA dataset, and as a result can be easily applied to aircraft data of vehicles with different specifications, such as non-autonomous aircraft. Valued future research into the real-time applications of this method as a supplement to current anomaly detection methods would provide an indication of the real-life functions of this system for aerospace and manufacturing applications. The versatility of the dynamic thresholding and weighted loss methods means that they can easily be applied on a variety of machine learning architectures, and they may provide similar enhancements as were found in this study.